\documentclass[letterpaper]{article} % DO NOT CHANGE THIS
\usepackage{aaai23}  % DO NOT CHANGE THIS
\usepackage{times}  % DO NOT CHANGE THIS
\usepackage{helvet}  % DO NOT CHANGE THIS
\usepackage{courier}  % DO NOT CHANGE THIS
\usepackage[hyphens]{url}  % DO NOT CHANGE THIS
\usepackage{graphicx} % DO NOT CHANGE THIS
\usepackage{natbib}  % DO NOT CHANGE THIS AND DO NOT ADD ANY OPTIONS TO IT
\usepackage{caption} % DO NOT CHANGE THIS AND DO NOT ADD ANY OPTIONS TO IT
\usepackage{algorithm}
\usepackage{algorithmic}
\usepackage{subfigure}
\usepackage{amsmath}
\usepackage[T1]{fontenc}
\usepackage{soul}
\usepackage{xcolor}
\usepackage{CJKutf8}
\usepackage{multirow}

\usepackage{newfloat}
\usepackage{listings}
\DeclareCaptionStyle{ruled}{labelfont=normalfont,labelsep=colon,strut=off} % DO NOT CHANGE THIS
\floatstyle{ruled}
\newfloat{listing}{tb}{lst}{}
\floatname{listing}{Listing}
\nocopyright 
\title{\fontsize{14}{16}PUnifiedNER: A Prompting-based Unified NER System for Diverse Datasets}
\author{
    Jinghui Lu\textsuperscript{\rm 1},
    Rui Zhao\textsuperscript{\rm 1},
    Brian Mac Namee\textsuperscript{\rm 2,3},
    Fei Tan\textsuperscript{\rm 1}\thanks{Corresponding author.}
}
\affiliations{
    \textsuperscript{\rm 1}SenseTime Research\\
    \textsuperscript{\rm 2}The Insight Centre for Data Analytics, University College Dublin\\
    \textsuperscript{\rm 3}School of Computer Science, University College Dublin\\
    \{lujinghui1, zhaorui, tanfei\}@sensetime.com, brian.macnamee@ucd.ie
}

\usepackage{bibentry}
\begin{document}
\begin{CJK}{UTF8}{gbsn}
\maketitle

\begin{abstract}
Much of named entity recognition (NER) research focuses on developing dataset-specific models based on data from the domain of interest, and a limited set of related entity types. This is frustrating as each new dataset requires a new model to be trained and stored. In this work, we present a ``versatile'' model---the \textbf{P}rompting-based \textbf{Unified NER} system (PUnifiedNER)---that works with data from different domains and can recognise up to 37 entity types simultaneously, and theoretically it could be as many as possible. By using prompt learning, PUnifiedNER is a novel approach that is able to jointly train across multiple corpora, implementing intelligent on-demand entity recognition. Experimental results show that PUnifiedNER leads to significant prediction benefits compared to dataset-specific models with impressively reduced model deployment costs. Furthermore, the performance of PUnifiedNER can achieve competitive or even better performance than state-of-the-art domain-specific methods for some datasets. We also perform comprehensive pilot and ablation studies to support in-depth analysis of each component in PUnifiedNER.
\end{abstract}

\section{Introduction}

% \tanfeicomment{}{1. abstract/contributions 2. novelty 3. demo/supplementary materials 4. figure 2 5. related works}

The Named Entity Recognition (NER) task involves the automatic recognition of entities in text with specific meaning, and so includes both entity extraction and entity classification. The recent rise of transformer-based, pre-trained language models \cite{devlin-etal-2019-bert,JMLR:v21:20-074,tan-etal-2020-tnt,tan-etal-2021-bert,lu-etal-2022-rationale,mao2022sda} has led to significant performance improvements in NER for various scenarios \cite{fries2022bigbio,parmar-etal-2022-boxbart}.

% Usually recognised named entities carry important information such as the names of people, places, or organisations as well as vocabulary from a domain of interest such as medical and legal terms.
% Therefore, NER provides the basis for many subsequent NLP tasks that require computable and inferable data including relationship extraction, syntax analysis, information retrieval, and QA systems.

% Figure \ref{fig:ner_demo} shows the use of NER in medical domains, which automatically extracts the keywords ``rectal cancer'', ``general anesthesia'' and ``transabdominal rectal cancer resection'' for a sentence, and classifies the entity types as ``disease'', ``anesthesia'' and ``surgery'' respectively. 

\begin{figure*}[!t]
\centering
\includegraphics[width=1.0\textwidth]{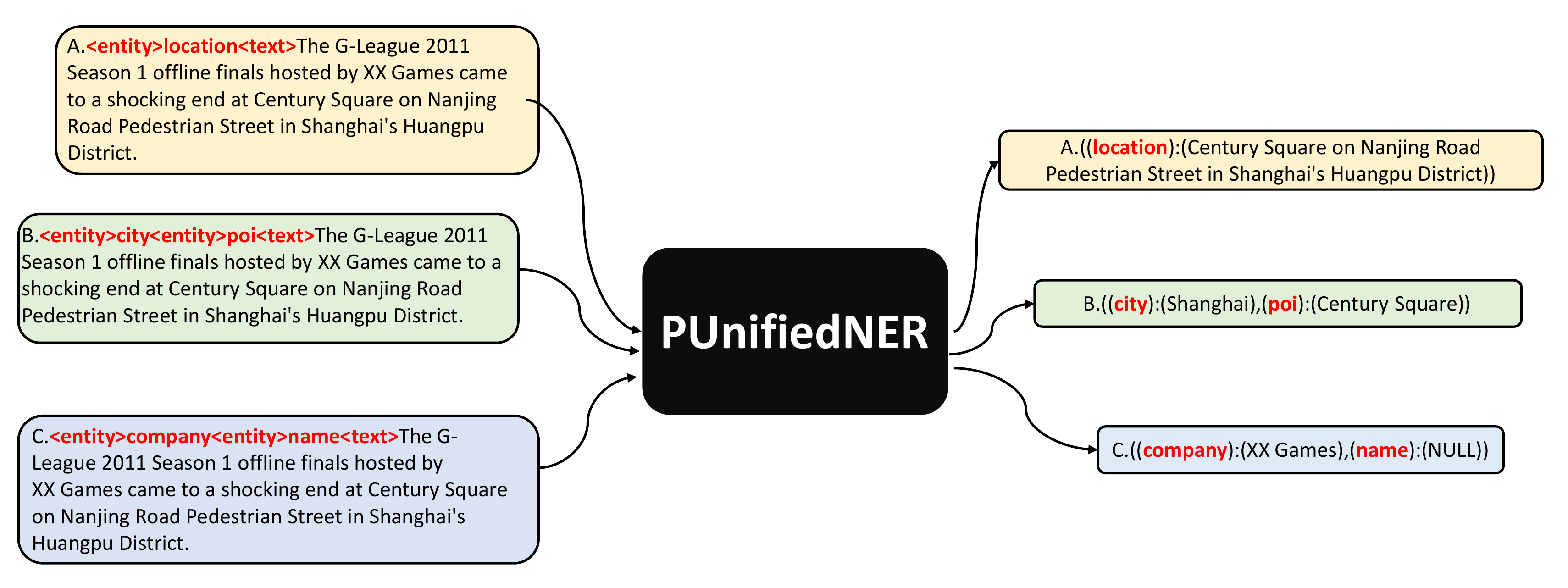}
 \caption{An overview of PUnifiedNER which reframes NER as a seq2seq task where red texts are prompts to suggest which entities users are interested in. For the same text input, PUnifiedNER returns different results conditioned on prefixed prompts.}\label{fig:overview}
\end{figure*}

Modern NER models perform self-supervised learning on large unlabelled text datasets to learn generic linguistic representations, and then are fine-tuned for a specific NER task on domain-specific datasets. 
The output of these approaches, however, is still a single dataset-specific model capable of recognising a few entity types, rather than a ``versatile'' model that can handle multiple domains and a large number of entity types simultaneously. This is unsatisfactory in practice: for example, an NER model trained on an e-commerce dataset can only extract entity types related to the e-commerce domain such as ``company'' and ``commodity'' but not named entities from other domains such as ``location'', ``organisation'', etc. As a result, such NER methods can not scale up well as each new dataset demands a new model to be trained and stored. 

We believe that it is more appealing to have a versatile model handle all scenarios. Besides, a unified model can achieve better performance than dataset-specific models if the unified model can be trained jointly, incorporating label information from different datasets. To be specific, although most NER datasets focus on corpora from various domains with different entity types, the underlying semantics or entity recognition are shared across corpora and exploiting this shared semantic information can enhance model robustness.

We hypothesise that this enhancement is derived from two aspects: commonality and diversity of label information. The enhancement from commonality refers to the fact that the model can see more relevant phrases from the same entity type. For example, several datasets with the ``location'' entity can capture most of the keywords belonging to the ``location'' entity in the real world. The enhancement from diversity means that the model can perceive more polysemous words. For example, ``apple'' could be a ``company'' entity (Apple Inc.) in a financial NER dataset, but may be a ``commodity'' entity (fruit apple) in an e-commerce dataset. Access to both senses during training will potentially strengthen the robustness of a model with regard to the polysemy problem. Moreover, when applying an NER model to specific text,  users are not always interested in all of the entity types embedded in the text. Existing NER models typically always output all entity types that they can recognise, which is not an ideal delivery mode. One on-demand model that flexibly recognises requested entity types should be attractive to many users. 

% We argue that by introducing a prompting-based mechanism, we can train a unified model that can address all the above mentioned challenges simultaneously. 

In this work, we develop a versatile model that can handle multiple NER datasets simultaneously, addressing all of the challenges mentioned above --- \textit{the \textbf{P}rompting-based \textbf{Unified NER} system (PUnifiedNER)}. Empowered by prompt learning, PUnifiedNER is built upon the recently proposed T5 language model \cite{JMLR:v21:20-074}, and an overview of the model is shown in Figure \ref{fig:overview}.
We jointly train the model on eight different datasets, supporting a total of 37 entity types. An obvious benefit of a prompting-based design is that the prompts can be served as instructions to guide the model to provide different outputs depending on the entity types of interest to the user, examples of which are shown in the red prompt text on the left panels of Figure \ref{fig:overview}. 

In addition to the ability to deal with various entity types and domains, our experimental results show that PUnifiedNER can achieve better performance than models of the same architecture trained using single datasets.
This is a promising result, as different NER datasets vary greatly in corpus domain, entity annotation, and scale, and it is a difficult challenge to have a single versatile model capable of handling multiple scenarios simultaneously. In other words, joint training on multiple datasets mutually benefits each other rather than degrades each other, which has been viewed as a notorious challenge \cite{lu202012,kamath2021mdetr,li2022grounded}. Furthermore, PUnifiedNER is on par with state-of-the-art methods on OntoNotes 4.0 dataset \cite{pradhan-etal-2013-towards} and achieves a new state-of-the-art result on the Resume dataset \cite{zhang-yang-2018-chinese}. To conclude, our contributions are as follows:

\begin{itemize}

  \item {We propose PUnifiedNER a new method that supports the extraction and recognition of up to 37 entity types. We also show that given different prompts, the model is able to generate entity types of interest to the user, enabling on-demand entity recognition (Figure \ref{fig:overview}).}
  
  \item {We demonstrate that a single PUnifiedNER model works well on eight publicly available Chinese NER datasets. This represents a reduction from 2.72 billion parameters (parameter calculation based on the BERT-Large model) to 220 million parameters compared to using a set of dataset-specific models.}
  
  \item {Experimental results show that PUnifiedNER achieves significantly better performance than dataset-specific models with the same architecture. On average, PUnifiedNER results in an improvement of 1.33 points against those dataset-specific models on eight datasets, while handling multiple datasets simultaneously. Furthermore, PUnifiedNER creates a new state-of-the-art result on the Resume dataset (f-score 96.65 → 97.18) and achieves a result comparable to the state-of-the-art on the OntoNotes 4.0 dataset (f-score 83.08 vs. 82.56).}
  
  \item {Comprehensive ablation studies and pilot studies are conducted to verify the effectiveness of each component in PUnifiedNER, including Language Model Adaptation, Dataset-Dependent Prompting, etc.}

\end{itemize}

\section{Related Work}

\subsubsection{Prompting-based Approaches for NLP:} The essence of prompt learning is making better use of pre-trained language model by adding additional ``hints'' \cite{liu2021pre,supernaturalinstructions}. Inspired by this, a large number of prompting methods are proposed in the literature to reformulate downstream tasks into pre-training ones to further leverage pre-trained language models. These prompting approaches can be categorised according to their model architectures and associated pre-training tasks, including methods based on bidirectional encoder-only models like BERT \cite{sun2021nsp,lu2022makes}, causal language decoder-only models like GPT-3 \cite{NEURIPS2020_1457c0d6,xie2021explanation}, and encoder-decoder models like T5 \cite{khashabi-etal-2020-unifiedqa,lester-etal-2021-power}.

% Also, these additional prompts can be either discrete natural language tokens \cite{sun2021nsp} or specific-designed continuous embeddings learned from data \cite{li-liang-2021-prefix,liu-etal-2022-p}.

There are also some studies exploring using prompting to unify various tasks, which is inline with our usage of prompting. However, they focus on various classification \cite{lester-etal-2021-power} or QA tasks \cite{khashabi-etal-2020-unifiedqa} instead of NER problems.

\subsubsection{Deep Learning Approaches for NER:} Most work considers NER as a sequence labelling task, where each token is assigned a pre-defined tag (e.g., BIO scheme). In this line of work, usually deep neural networks such as bidirectional LSTM \cite{zhang-yang-2018-chinese,liu-etal-2019-encoding} or pre-trained transformer-based language models \cite{ma-etal-2020-simplify,liu-etal-2021-lexicon} is combined with Conditional Random Fields (CRF) \cite{lafferty2001conditional} and has been widely used in solving \textit{flat NER}, where neither overlapped entities nor non-adjacent entities will appear in the text.

Inspired by the success of using deep learning model in flat NER task, many work in the literature attempts to solve \textit{nested NER} (i.e., entities have overlaps with each other) or \textit{discontinuous NER} (i.e., entities consist of non-consecutive text sequence) by reformulating the token-level sequence labelling methods used in flat NER to span-level sequence labelling methods \cite{katiyar-cardie-2018-nested,yu-etal-2020-named,shen-etal-2021-locate,li-etal-2021-span}, where text spans instead of tokens are 
enumerated and classified. Seq2Seq model is also{applicable for NER, which takes as input a sentence and generates a sequence of entity offsets, span lengths as well as labels \cite{fei2021rethinking}. 

There is an important research thread attempting to unify three sub-tasks in NER into one framework. \citet{li2022unified} propose W\textsuperscript{2}NER framework that reframes NER as a word-word relation classification problem and reach the state-of-the-art performance in 14 NER datasets. \citet{yan-etal-2021-unified-generative} apply seq2seq model to generate a sequence of entity start-end indexes and types. More recently, \citet{lu-etal-2022-unified} propose a text-to-structure framework based on seq2seq model to unify not only NER tasks but other information extraction tasks (i.e., event, relation and sentiment extraction).

\subsubsection{Differences between PUnifiedNER and Other Unified Approaches:}
Though unifying various sub-tasks, most of existing work essentially tries to build up a dedicated model for every single dataset. The most significant difference is that, we focus on training a versatile model that handles a large scale of entity types as well as datasets simultaneously, which is more compelling.

Other differences are (1) PUnifiedNER vs. W\textsuperscript{2}NER: we reframe NER as seq2seq while W\textsuperscript{2}NER recasts NER as word-word relation classification, which is totally different, and PUnifiedNER outperforms W\textsuperscript{2}NER in dataset Resume. (2) PUnifiedNER vs. other seq2seq-based NER approaches: most seq2seq methods still consider identifying more accurate entity boundaries, thus, the output is entity offsets and span lengths, or entity start-end indexes. PUnifiedNER elegantly generates entity types and corresponding text spans using natural language, which also indicates the potential of tackling nested and discontinuous NER. We leave these explorations for future work. Besides, enhanced by prompt-learning, our model implements the on-demand named entity recognition which has not been fully explored in existing studies.

% Also, our scope is unifying all datasets within flat NER instead of unifying three NER sub-tasks. In spite of this, PUnifiedNER produces natural language as output 

\section{PUnifiedNER}

% Our unified NER system is based on the recent seq2seq frameworks, i.e., T5 \cite{JMLR:v21:20-074} (Section \nameref{subsec:model_archi}).\footnote{We use the T5-v1.1-base-chinese checkpoint pre-trained by UER: https://huggingface.co/uer/t5-base-chinese-cluecorpussmall.} We then introduce how we reframe NER task of different datasets into a unified seq2seq format with prompts (Section \ref{subsec:datareframe}). Then we present several datasets used in our experiments (\tanfeicomment{}{Section} \ref{subsec:datasets}). We then introduce how we train PUnifiedNER that is a NER system applicable on large scale named entities indicating comparable results compared to \tanfei{}{the} state-of-the-art \tanfei{}{approaches} on several datasets (Section \ref{subsec:lmadaptation} and \ref{subsec:mdtraining}).

Our unified NER system is based on the recent seq2seq frameworks, i.e., T5 \cite{JMLR:v21:20-074}.\footnote{We use the T5-v1.1-base-chinese checkpoint pre-trained by UER: \url{https://huggingface.co/uer/t5-base-chinese-cluecorpussmall}.} We first present the model architecture and describe how we reframe the NER task on different datasets into a unified seq2seq format with prompts. We then present the datasets used in our experiments. Finally, we detail the way PUnifiedNER is trained.

\subsection{Model Architecture}\label{subsec:model_archi}

% \begin{figure}[!t]
% \centering
% \includegraphics[width=.5\textwidth]{figures/t5.png}
%  \caption{T5 architecture overview.\protect\footnotemark}
%  \label{fig:t5}
% \end{figure}

% \footnotetext{The figure is taken from http://jalammar.github.io/illustrated-transformer/.}

Our model is mainly implemented by reusing the pre-trained T5 model. T5 follows the vanilla transformer encoder-decoder architecture \cite{vaswani2017attention} with some minor modifications, described in \citet{JMLR:v21:20-074}. The encoder, consisting of multiple transformer encoder blocks, uses the bidirectional multi-head self-attention mechanism to encode the input text sequence. The decoder, consisting of multiple transformer decoder blocks, receives the output hidden state from the encoder and uses the unidirectional attention mechanism to autoregressively generate the output text sequence. The predicted token generated at each step is compared to the ground truth token, supervised using a cross entropy loss.

\subsection{Task Reframing}\label{subsec:datareframe}

As shown in Figure \ref{fig:overview}, we reframe NER as a prompting-based seq2seq problem. Formally, given an original text sequence $x$, we transform $x$ to a source sequence $x_{input}$ by prefixing it with a series of prompts as follows:

\begin{equation} \label{eq:prefix_x}
\begin{split}
x_{input} & = [s_{e},s_{p_{1}},...,s_{e},s_{p_{n}},s_{t},x]
\end{split}
\end{equation}

\noindent where $x_{input}$ is the model input; $s_{e}$ is the special token \textit{``<entity>''} indicating that the following token is the entity type that we are interested in; $s_{p_{i}}$ is the entity type, e.g., ``city''; and $s_{t}$ is the special token \textit{``<text>''} indicating that the following text sequence is the sentence from which entities should be extracted. Then the target sequence $y_{output}$ is as follows:

\begin{equation} \label{eq:prefix_y}
\begin{split}
y_{output} & = ((s_{p_{1}}):(ent_{1}),...,(s_{p_{n}}):(ent_{n}))
\end{split}
\end{equation}

\noindent where $y_{output}$ is the model output; $s_{p_{i}}$ are entity types that are identical to those in Equation \ref{eq:prefix_x}; and $ent_{i}$ is the ground truth texts extracted from the input sequence. 

% For example, suppose we have an input $x$ ``李明明天去植物园。'' and we are interested in entities <时间 (time)> and <地点 (location)>. Then the $x_{input}$ will be ``<实体><时间><实体><地点><文本>李明明天去植物园。'' and the $y_{output}$ will be ``((时间):(明天),(地点):(植物园))''. Alternatively, if we are interested in entities <名称 (name)>, then the $x_{input}$ will be ``<实体><名称><文本>李明明天去植物园。'' and the $y_{output}$ will be ``((名称):(李明))''. In this case, we hope prompts can serve as indicators that suggest which entities users are interested in to steer model to produce different outputs even given the same input $x$ (also see example A and B in Figure \ref{fig:overview}). Noted that if users assign an entity not in input $x$, the $ent_{i}$ in $y_{output}$ will be a special token ``NULL'' (see example C in Figure \ref{fig:overview}).

For example, suppose we have an input $x$ \textit{``Tom will go to the zoo tomorrow.''} and we are interested in entities ``time'' and ``location''. Then $x_{input}$ will be \textit{``<entity><time><entity><location><text>Tom will go to the zoo tomorrow.''} and $y_{output}$ will be \textit{``((time):(tomorrow),(location):(zoo))''}. Alternatively, if we intend to parse entities ``name'', then  $x_{input}$ will be \textit{``<entity><name><text>Tom will go to the zoo tomorrow.''} and $y_{output}$ will be \textit{``((name):(Tom))''}. In this case, we hope prompts can function well as indicators that suggest which entities users are interested in to steer model produce different entities, even given the same input $x$ (also see example A and B in Figure \ref{fig:overview}). Note that if users assign an entity absent in input $x$, the $ent_{i}$ in $y_{output}$ will be a special token \textit{``NULL''} (see example C in Figure \ref{fig:overview}).

We argue that this design has several advantages: (1) prompts serve as indicators of target entities as discussed above; (2) unifying entities from different datasets into a single model reduces the cost of model deployment compared to using a dedicated model for each dataset; and (3) reusing labelled information of different datasets. Many NER datasets share similar entity types such as ``name'', ``location'' and ``organisation''. Training a model solely on a single dataset can only learn knowledge of the given dataset. PUnifiedNER can  exploit the shared knowledge crossing diverse datasets by using prefixed prompting. 

% (4) Potential of solving nested NER and discontinuous NER though it is not the scope of this work.

\subsection{Datasets} \label{subsec:datasets}

We train and evaluate PUnifiedNER on eight existing public NER datasets that target various entity types from different domains including social media, e-commerce, news, postal address, etc. We use the \textit{Ecommerce} \cite{ding2019neural}, \textit{MSRA} \cite{levow-2006-third}, \textit{OntoNotes 4.0} \cite{pradhan-etal-2013-towards}, \textit{People Daily 2014}, \textit{Boson},\footnote{People Daily 2014 and Boson datasets are available at https://github.com/hspuppy/hugbert/tree/master/ner\_dataset.} \textit{Resume} \cite{zhang-yang-2018-chinese}, \textit{CCKS2021},\footnote{https://tianchi.aliyun.com/competition/entrance/531900/infor-mation} and \textit{CLUENER} \cite{xu2020cluener2020} datasets. After preprocessing, these datasets provide 37 unique entities. Figure \ref{fig:ner_tag_summary} summarises the properties of these entities. They are categorised into four big groups---name, location, organisation, and other---based on their semantic information and three granularities---coarse-grained, fine-grained, and ultra fine-grained. Note that all texts are originally in Chinese, so we provide translations in this paper to facilitate understanding. Additional details of the datasets and entities are provided in Appendix.\looseness=-1

%  \ref{sec:appendix}.

\begin{figure}[!t]
\centering
\includegraphics[width=.5\textwidth]{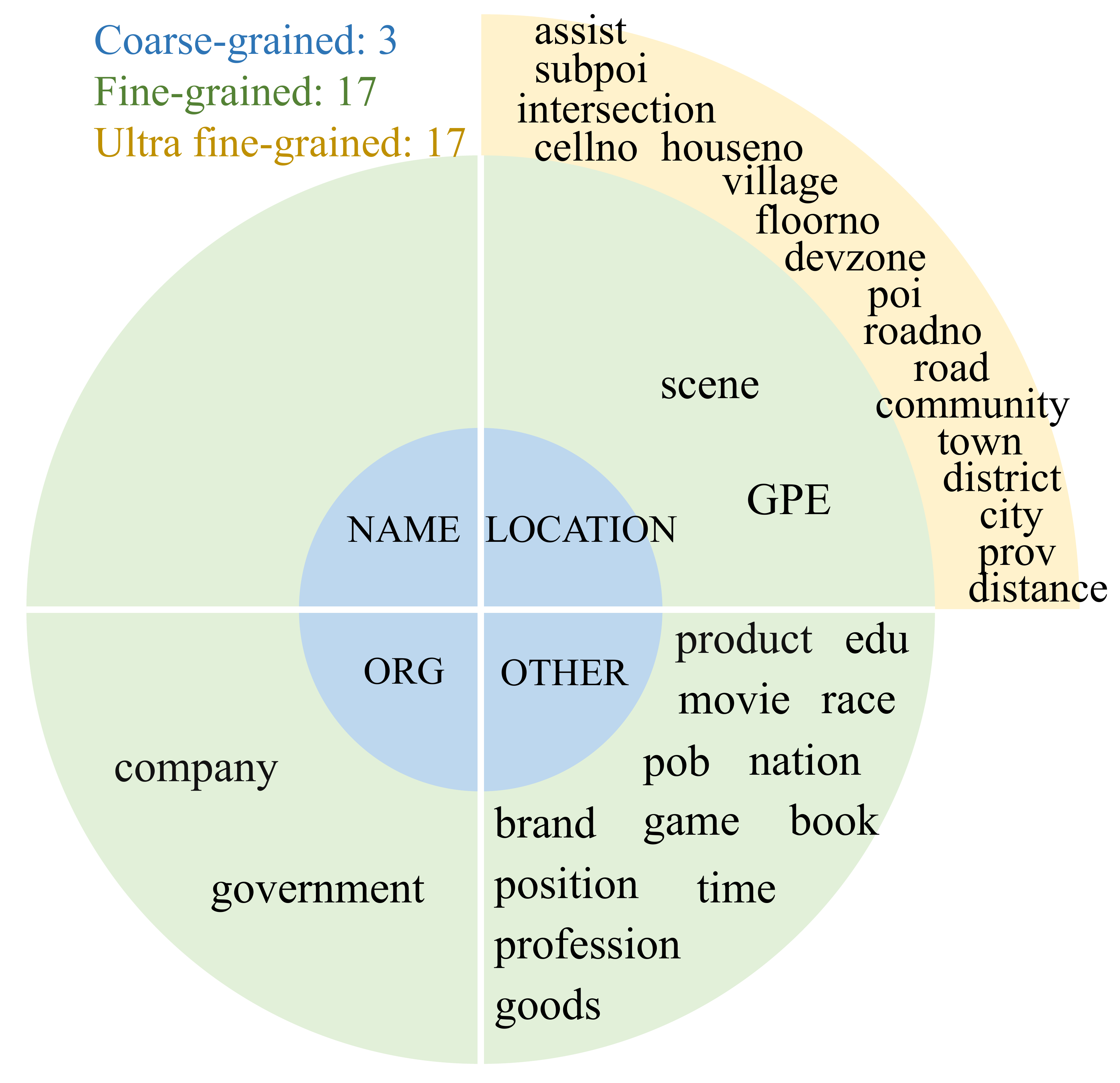}
 \caption{Properties of 37 entities provided by eight public NER datasets. All entities are manually categorised into four groups, i.e., name, location, organisation, and others. Entities in the same quadrant fall into the same group. Entities in different colors indicate different entity granularity, i.e., blue (coarse-grained), green (fine-grained) and yellow (ultra fine-grained). Note that the manually crafted granularity/category information in this figure is for a better visualisation and is not leveraged in model training.}\label{fig:ner_tag_summary}
\end{figure}

\subsection{Language Model Adaptation} \label{subsec:lmadaptation}

T5 is pre-trained with ``reconstructing'' masked spans in the source sequence, which are marked with unique sentinel tokens. The target output sequence consists of several sentinels that are followed by the corresponding masked content. Concretely, given a text \textit{``The capital of China is Beijing.''}, the source sequence of the pre-trained example might be constructed as \textit{``The <extra0> of <extra1> is Beijing.''} and the target sequence will be \textit{``<extra0>capital<extra1>China<extra2>''} where \textit{``<extra\textsubscript{i}>''} are sentinels and the last sentinel (\textit{``<extra2>''} in this case) is the end of sequence token.

Although this pre-training objective works effectively in \citet{JMLR:v21:20-074}, we believe that this setup is not good enough to shift the pre-trained model to our downstream prompting-based seq2seq NER task. One main obstacle is the objective gap where the pre-trained T5 has never seen a natural and complete input sequence (i.e., text sequence without sentinels) that is crucial for NER tasks. The NER model needs to extract text spans correctly from the original input given the completed context. Therefore, inspired by the idea of continuously pre-training language models \citep{khashabi-etal-2020-unifiedqa,lu202012,lester-etal-2021-power,lu-etal-2022-unified}, we adopt the prefix language modelling objective discussed by \citet{JMLR:v21:20-074,lester-etal-2021-power} to further adapt the pre-trained language model: we randomly split a given natural text into two substrings and the model must produce the latter substring conditioned on the former substring.  For example, given a original sentence \textit{``The capital of China is Beijing.''}, the source input of a pre-trained example might be constructed as \textit{``The capital of China''} and the target output will be \textit{``is Beijing.''}, respectively.\looseness=-1

% For example, given the sentence \textit{``The capital of China is Beijing.''} as source input, a target output might be \textit{``The capital of China''} and \textit{``is Beijing.''}, respectively. 

We conduct prefix language modelling training based on T5-v1.1-base-chinese over all NER datasets, encouraging the model to close both the objective gap and the domain gap simultaneously. We have found that training using this self-supervised objective can largely close the objective and domain gaps as well as boosting model performance. This is presented in the pilot study. \looseness=-1

\subsection{Multi-Dataset Joint Learning} \label{subsec:mdtraining}

After language model adaptation we adopt a simple, yet effective, multi-dataset learning strategy to train the model: for each batch, we randomly select examples from different datasets until the number of examples is identical to the batch size.

Note that during multi-dataset training, it is unfeasible to use all entities as prefixed prompts---prefixing all entity prompts results in around 296 extra tokens (avg. 8 for each entity in Chinese) to the input sequence as well as multiple \textit{``NULL''} tokens in the target output sequence. Therefore, we propose three prefixed prompt setups: 

\begin{itemize}

\item \noindent \textbf{Random Prompt:} During training, we randomly sample up to all 37 entities as prefixed prompts. While during inference, we use all entities of the specific testing dataset as prefixed prompts. For example, the MSRA dataset contains three entities---location, organisation, and name---so during inference the prefixed prompts are always \textit{``<entity>-<location><entity><organisation><entity><name><te-xt>''}.

\item \noindent \textbf{Random Prompt + Exact Match:} During training, each original example is prefixed in two styles (therefore two training examples are generated), one using Random Prompt as discussed above, and the other using the exact entities from the ground truth as prefixed prompts (i.e., Exact Match). During inference, we use all entities of the specific testing dataset as prefixed prompts. Though this setting suffers from information leakage where ground truth entities can help the model narrow down the entity space to reduce difficulty during training, we do not use Exact Match during inference. Thus it is still a fair comparison. Also, we hope the use of Random Prompt can partially alleviate the information leakage problem.  

\item \noindent \textbf{Dataset-Dependent Prompt:} Training and inference both use all entities from the specific dataset as prefixed prompts. 

\end{itemize} 

The effectiveness of these three prefixed prompt settings is explored in the ablation study.

\section{Pilot Study: The Benefits of Language Model Adaptation} \label{sec:lmadaptation}

We first ask a  question: \textit{Is it necessary to apply language model adaptation before multi-dataset learning?} To answer this question, we first train a T5-v1.1-base-chinese checkpoint on the CLUENER dataset using the Dataset-Dependent Prompt approach discussed in previous section. The best validation f-score achieved is 48.67 which is far from ideal.\looseness=-1 

Then we continuously pre-train T5 model with the prefix language modelling objective over eight public datasets. The training/validation split is the same as the original setting. We also use the sampling strategy discussed in the previous section to construct training batches and continuously pre-train T5 up to 50K steps, evaluating every 1K steps (other training details are provided in the Appendix). Figure \ref{fig:prefix_lm} shows the prefix language modelling validation loss for the eight datasets. We find that for most datasets, the model achieves the lowest validation loss with at least 5K steps.

% Given that we use batch size of 128 and the model still costs thousands of steps to converge, it is not a trivial task to close objective gap.

Finally, we select the model after 6K steps, which has on average a lower validation loss across all eight datasets, and retrain it with the CLUENER dataset. The best validation f-score is 74.78, outperforming the previous f-score of 48.67 by a large margin. This demonstrates the necessity and effectiveness of applying language model adaptation to close the gap between T5 pre-training and downstream prompting-based seq2seq NER tasks. Given the benefit brought by language model adaptation, we will use the model at prefix language modelling step 6K as our backbone model unless otherwise stated.\looseness=-1

% \begin{figure}[!t]
%     \centering
%     \subfigure[]{\includegraphics[width=0.23\textwidth]{figures/cluener_eval_loss.pdf}} 
%     \subfigure[]{\includegraphics[width=0.23\textwidth]{figures/msra_eval_loss.pdf}} 
%     \subfigure[]{\includegraphics[width=0.23\textwidth]{figures/resume_eval_loss.pdf}\label{subfig:resume}}
%     \subfigure[]{\includegraphics[width=0.23\textwidth]{figures/ontonote_eval_loss.pdf}}
%     \subfigure[]{\includegraphics[width=0.23\textwidth]{figures/chinese_address_eval_loss.pdf}}
%     \subfigure[]{\includegraphics[width=0.23\textwidth]{figures/pd_eval_loss.pdf}} 
%     \subfigure[]{\includegraphics[width=0.23\textwidth]{figures/boson_eval_loss.pdf}}
%     \subfigure[]{\includegraphics[width=0.23\textwidth]{figures/ecommerce_eval_loss.pdf}}
%     \caption{Prefix language modelling validation loss of (a) CLUENER (b) MSRA (c) Resume (d) OntoNotes 4.0 (e) CCKS2021 (f) People Daily 2014 (g) Boson (h) Ecommerce.}
%     \label{}
% \end{figure}

\begin{figure}[!t]
\centering
\includegraphics[width=.5\textwidth]{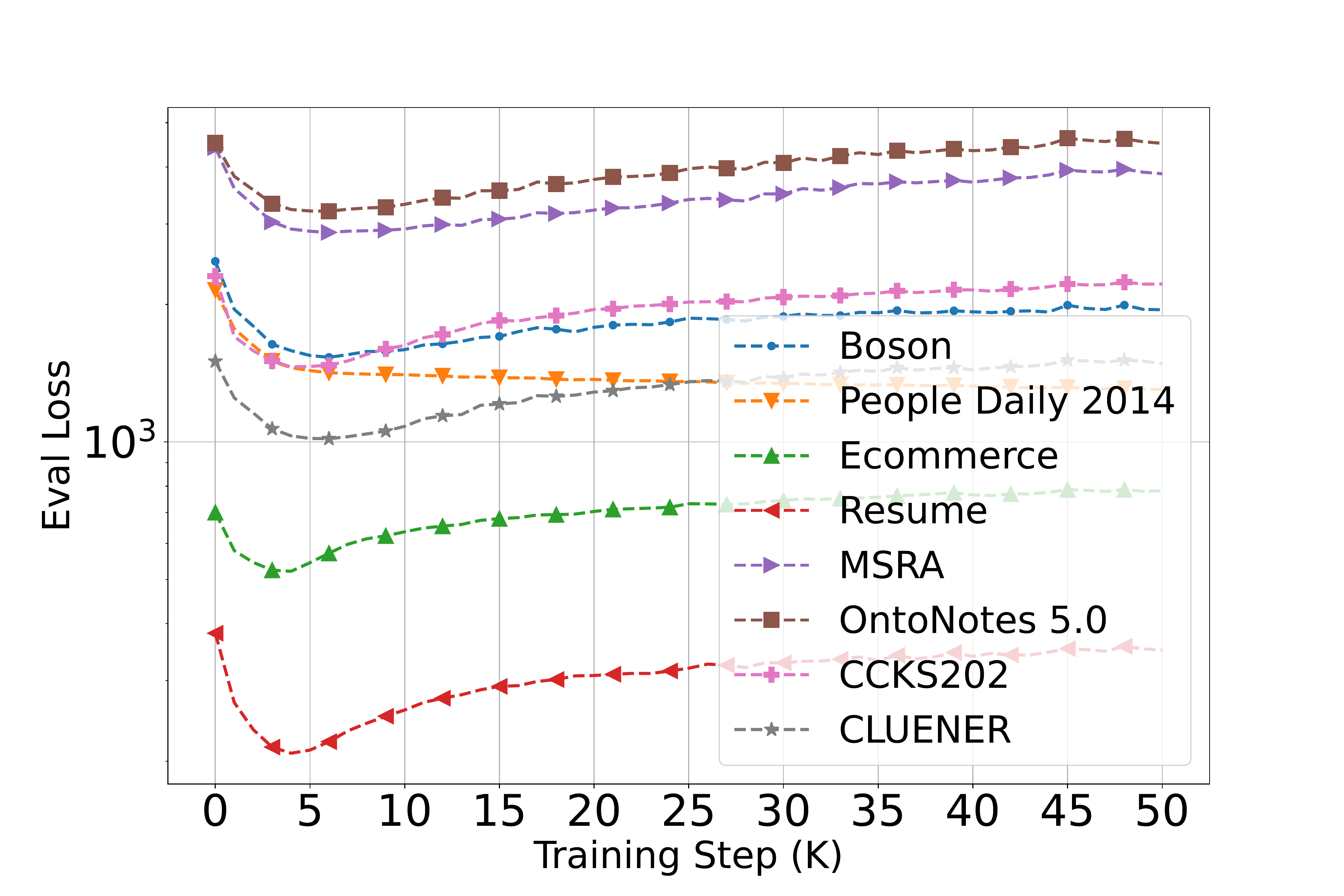}
 \caption{Prefix language modelling validation loss on eight datasets. Loss is log scaled to make trends more noticeable.}
 \label{fig:prefix_lm}
\end{figure}

\section{Experiments}

\begin{table*}[!htb]
\small
\centering

\begin{tabular}{llllllllllll}
\hline
\textbf{Methods}                 & \textbf{MSRA*}       & \textbf{OntoNotes} & \textbf{Resume} & \textbf{CLUENER}   & \textbf{Ecom}      & \textbf{PD} & \textbf{Boson} & \textbf{CCKS*} & \textbf{Avg.}& \textbf{\# models} \\ \hline
Single-Dataset& \textbf{87.78} & 78.44   & 93.91      & 74.57          & 69.38              & \textbf{96.99}            & 82.60     & 84.26        & 83.49 &8         \\
PUnifiedNER & 87.07      & \textbf{82.56}         & \textbf{97.18}  & \textbf{77.00}      & \textbf{71.32} & 94.49        & \textbf{83.58} & \textbf{85.34}    & \textbf{84.82} &1         \\\hline
% On-demand Fine-Tuning   & 88.32*          & 64.54*             & 96.60*      & 73.64* & 71.40*              & 98.44*            & 78.78*     & 86.31*        & 8         \\ \hline
\end{tabular}%

\caption{Comparison of PUnifiedNER to single-dataset training. ``*'' indicates that this dataset does not contain the test set, thus we report the results of validation set.}
\label{tab:vs_single_demand}
\end{table*}

\begin{table*}[!t]
\small
\centering
\resizebox{.9\textwidth}{!}{%
\begin{tabular}{lllllll}
\hline
\textbf{Flat NER Methods} & \textbf{MSRA}          & \textbf{OntoNotes}  & \textbf{Resume}         & \textbf{CLUENER}        & \textbf{Ecommerce}     & \textbf{\# Params} \\ \hline
Lattice LSTM \cite{zhang-yang-2018-chinese}            & 93.18             & 73.88              & 94.46              & -              & - & -         \\
TENER \cite{yan2019tener}            & 92.74             & 72.43              & 95.00              & -              & - & -         \\
LGN \cite{gui-etal-2019-lexicon}            & 93.71             & 74.45              & 95.11              & -              & - & -         \\
FLAT \cite{li-etal-2020-flat}            & 96.09             & 81.82              & 95.86              & -              & - & N*110M         \\
SoftLexicon \cite{ma-etal-2020-simplify}            & 95.42             & 82.81              & 96.11              & -              & - & N*110M             \\ 
LEBERT \cite{liu-etal-2021-lexicon}            & 95.70             & 82.08              & 96.08              & -              & - & N*110M             \\ \hline
\textbf{State-of-the-art Methods}                   &        &        &  &          &      &    \\ \hline
W\textsuperscript{2}NER \cite{li2021unified}               & \textbf{96.40} & \textbf{83.08} & 96.65          & -              & -             & N*110M       \\
NEZHA-BC \cite{zhang-etal-2022-domain}            & -             & -              & -              & -              & \textbf{78.69} & N*110M          \\
DML-NEZHA-Large               & -             & -              & -              & \textbf{83.30} & -             & N*340M          \\ \hline
\textbf{Our Method}                   &        &        &  &          &      &    \\ \hline
PUnifiedNER                    & 87.07          & 82.56          & \textbf{97.18} & 77.00          & 71.32         & 1*220M        \\ \hline
\end{tabular}
}
\caption{Comparison of the performance of PUnifiedNER to recent state-of-the-art methods. All state-of-the-art methods are dataset-specific models. Estimated number of parameters used for model deployment in realistic scenarios are shown in the rightmost column, where $N$ is the number of datasets. The number of parameters for some methods are estimated from the corresponding paper.}
\label{tab:vs_stoa}
\end{table*}

\begin{table*}[!t]
\small
\centering
\begin{tabular}{lllllll}
\hline
\textbf{Prompt}                      & \textbf{MSRA}           & \textbf{OntoNotes}  & \textbf{Resume}         & \textbf{CLUENER}        & \textbf{CCKS2021}   & \textbf{Avg. Score}    \\ \hline
Random Prompt               & 69.93          & 46.88          & 95.48          & 69.62          & 84.30   & 73.24       \\
Random Prompt + Exact Match & 69.21          & 55.47          & 90.20           & 66.30           & 80.37   & 72.31       \\
Dataset-Dependent Prompt     & \textbf{87.07} & \textbf{82.56} & \textbf{97.18} & \textbf{77.00} & \textbf{85.34} & \textbf{85.83} \\ \hline
\end{tabular}%

\caption{Ablations on our Prefixed Prompts design choices.}
\label{tab:ablation_prefixed_design}
\end{table*}

\begin{table*}[!t]
\small
\centering

\begin{tabular}{llllllllll}
\hline
\textbf{Step}   & \textbf{MSRA}           & \textbf{OntoNotes}  & \textbf{Resume}         & \textbf{CLUENER}        & \textbf{CCKS}       & \textbf{Ecom}      & \textbf{People Daily}   & \textbf{Boson} & \textbf{Avg. Score}         \\ \hline
Step 6K  & \textbf{87.07} & \textbf{82.56} & 97.18          & \textbf{77.00} & 85.34          & \textbf{71.32} & 94.49          & \textbf{83.58} & \textbf{84.82} \\
Step 10K & 86.94          & 82.09          & 96.62          & 76.16          & \textbf{86.01} & 71.23          & 95.27          & 79.69      & 84.25    \\
Step 50K & 86.64          & 81.86          & \textbf{97.24} & 75.21         & 84.83          & 70.52          & \textbf{96.95} & 82.84   & 84.51       \\ \hline
\end{tabular}%

\caption{Ablations on language model adaptation steps.}
\label{tab:ablation_lm_steps}
\end{table*}

\subsection{Comparison to Single Dataset Performance} 

\begin{figure}[!t]
    \centering
    \subfigure[]{\includegraphics[width=0.47\textwidth]{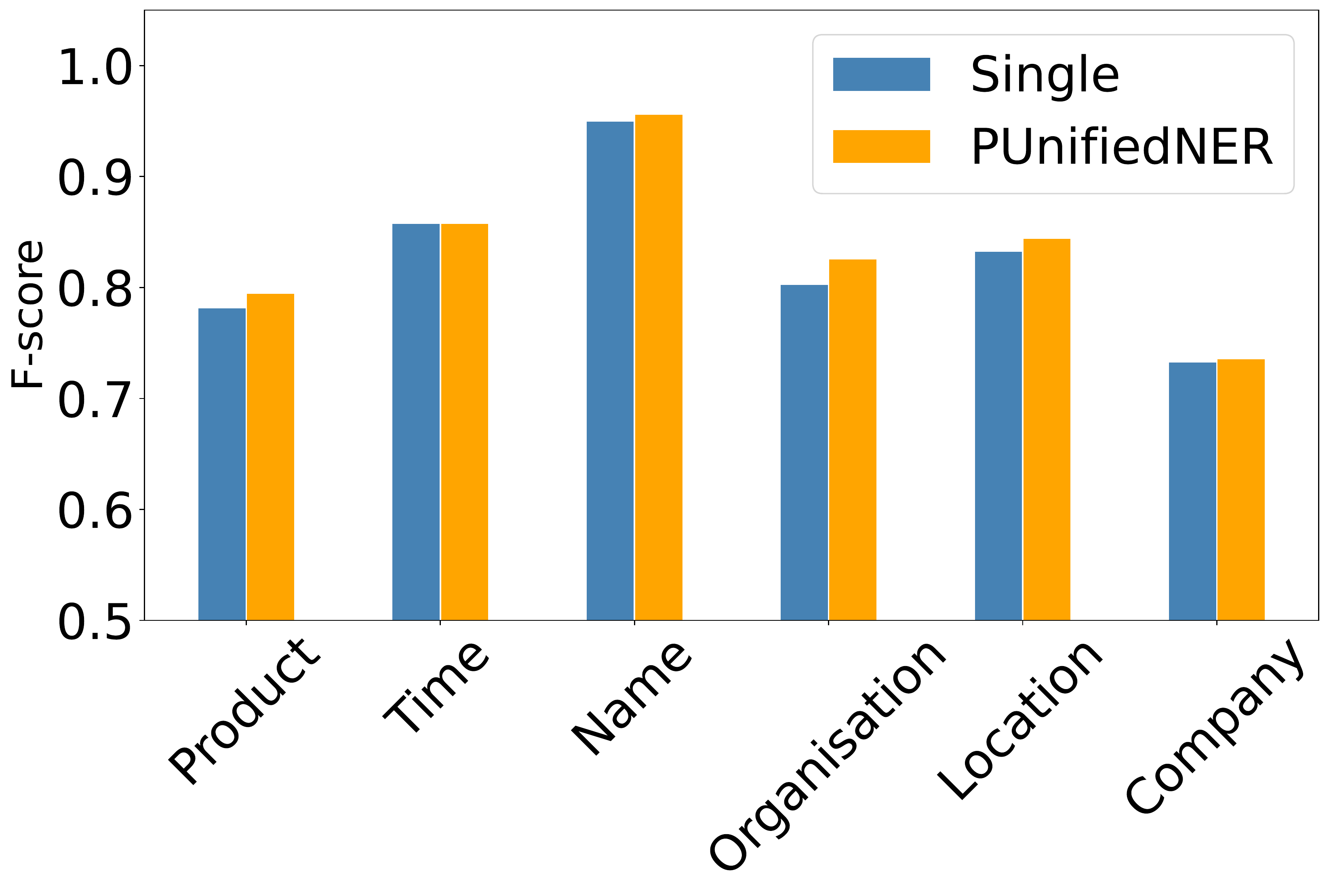}} 
    \subfigure[]{\includegraphics[width=0.47\textwidth]{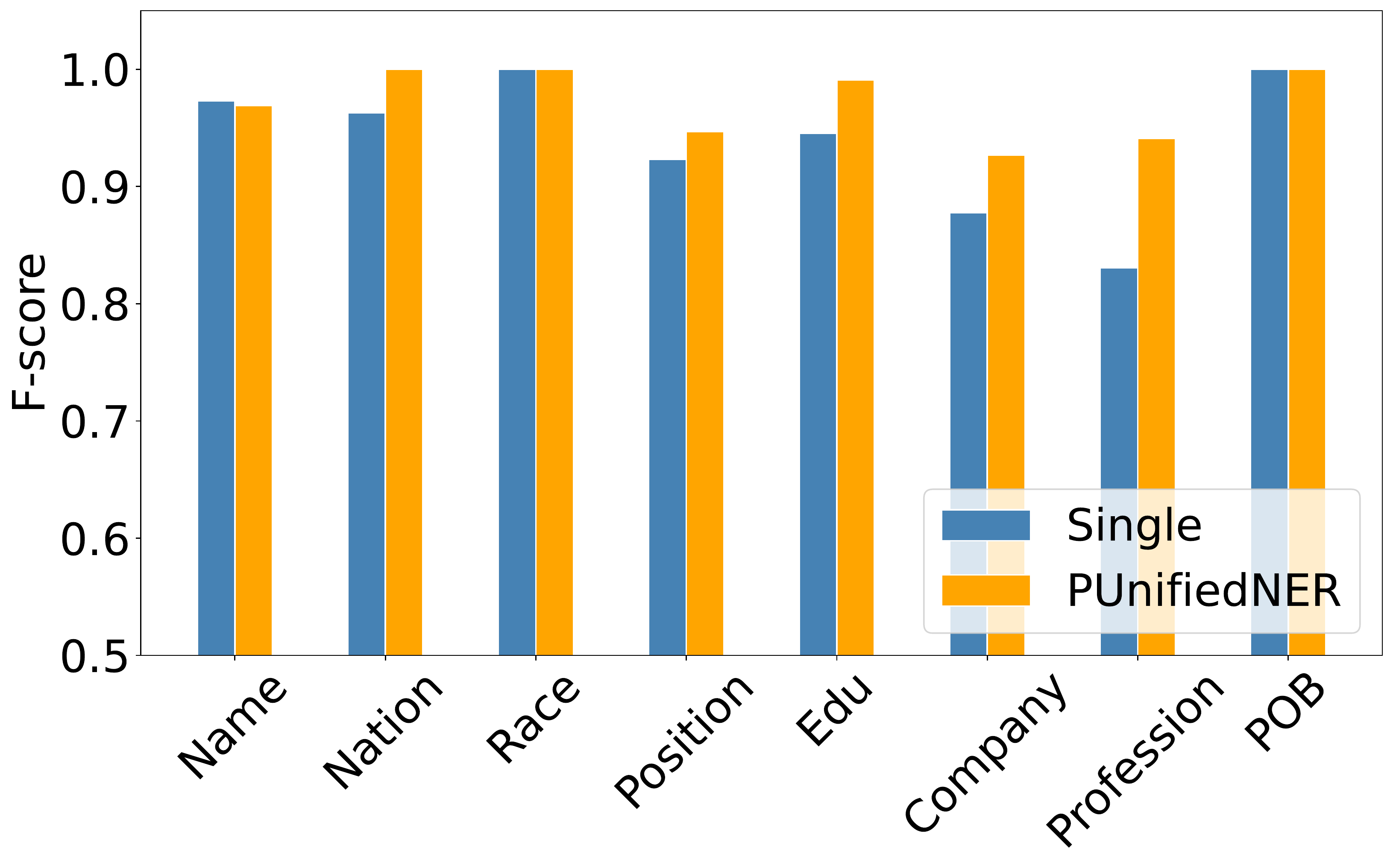}\label{subfig:fine-grained_resume}} 
    \caption{Fine-grained named entity f-score of (a) Boson (b) Resume where Edu and POB denote education and place of birth, respectively.}
    \label{fig:fine-grained}
\end{figure}

Our intention is to answer the question: \textit{Can a jointly trained PUnifiedNER model} learn shared information among datasets and hence outperform seq2seq NER models trained on single datasets?}

As with the pilot study, we first establish baseline performance of single-dataset training by fine-tuning the T5 backbone on each NER dataset independently. The fine-tuning procedure is the same as discussed in the pilot study. Then we jointly train another T5 backbone on multiple NER datasets using Dataset-Dependent Prompt settings and the same sampling strategy to construct batches described in the pilot study. For single-dataset training, we select the best performing model on a validation set and report its test f-score. For multi-dataset training, we select the model with best mean f-score on all eight NER datasets and report its test f-scores on each dataset separately. Other hyper-parameters settings are detailed in the Appendix. 

Experimental results are shown in Table \ref{tab:vs_single_demand}. It shows a clear pattern that the jointly trained PUnifiedNER model can learn shared information between datasets, surpassing single-dataset trained models by an average improvement of 1.33 points. To be specific, in six out of eight datasets, PUnifiedNER exceeds its single-dataset trained counterparts by a large margin especially in OntoNotes 4.0 (82.56 vs. 78.44), Resume (97.18 vs. 93.91) and CLUENER (77.00 vs. 74.57), and with a reasonable improvement in Ecommerce (71.32 vs. 69.38), Boson (83.58 vs 82.60) and CCKS2021 (85.34 vs. 84.26). In MSRA, the performance of PUnifiedNER is also comparable. The only exception is the People Daily 2014 dataset where the single-dataset trained model exceeds PUnifiedNER by 2.5 points. Given the fact that the People Daily 2014 dataset includes more than 400K instances, we believe that PUnifiedNER suffers from insufficient training regarding the People Daily 2014 dataset. This is probably because we select the checkpoint that is generally useful for all datasets according to performance on validation sets, where models are converged in all datasets other than People Daily 2014. In our follow-up observations, we found that the validation set performance of the PUnifiedNER model was still improving after the ``best'' checkpoint, and more details can be seen in the Appendix. We suspect that this can be alleviated by some multi-training strategies such as Curriculum Learning and Dynamic Stop-and-Go \cite{lu202012}, which we will address in future work.

\subsubsection{Fine-grained Analysis:} We further report fine-grained named entity f-scores for the Boson and Resume datasets in Figure \ref{fig:fine-grained}. First, for the Boson dataset the f-score achieved by PUnifiedNER for almost all entity types is improved compared to the corresponding single-training counterpart. This is consistent to our expectation because all entity types in Boson have appeared in other datasets and joint training can take advantage of ``seeing'' more data. It is surprising to find out that the prompting-based joint training can also bring additional discriminative ability for entity types that only appear in one dataset, which is evidenced by the f-scores for the Education and Profession entities in the Resume dataset as shown in Figure \ref{subfig:fine-grained_resume}. 

% \subsection{on-demand Fine-tuning}

% In some scenarios, single dataset performance is paramount while model size is not a concern. So we propose the question \textit{Whether we can improve the single dataset performance of multi-dataset trained PUnifiedNER through continuously Dataset-Dependent fine-tuning?} To address this question, we fine-tune the multi-dataset trained PUnifiedNER on eight datasets. Training settings are identical to those discussed in Section \ref{sec:lmadaptation} and hyper-parameters settings are shown in in Appendix \ref{sec:appendix}.

% Table \ref{tab:vs_single_demand} demonstrates ...

\subsection{Comparison to State-of-the-art Approaches}

In Table \ref{tab:vs_stoa} we compare the performance of  PUnifiedNER with recent state-of-the-art approaches, as well as other flat NER approaches including \textbf{Lattice LSTM} \cite{zhang-yang-2018-chinese}, \textbf{TENER} \cite{yan2019tener}, \textbf{LGN} \cite{gui-etal-2019-lexicon}, \textbf{FLAT} \cite{li-etal-2020-flat}, \textbf{SoftLexicon} \cite{ma-etal-2020-simplify}, and \textbf{LEBERT} \cite{liu-etal-2021-lexicon}. We draw special comparison with the recent \textbf{W\textsuperscript{2}NER} approach \citep{li2021unified} as it demonstrates very strong performance on several Chinese NER datasets. We also compare PUnifiedNER with \textbf{DML-NEZHA-Large}\footnote{State-of-the-art performance is reported at \url{https://www.cluebenchmarks.com/ner.html} as of July 1st, 2022.} and \textbf{NEZHA-BC} \cite{zhang-etal-2022-domain} since they achieve state-of-the-art results for the CLUENER and Ecommerce datasets, respectively. Our single multi-dataset jointly trained model achieves competitive performance with these state-of-the-art dataset-specific models in two out of five datasets. In particular, on the Resume dataset, PUnifiedNER surpasses the state-of-the-art (96.65 -> 97.18). 

Though the performance of PUnifiedNER is less satisfactory for some datasets e.g., MSRA, we should keep in mind that PUnifiedNER is a single model for diverse datasets while the other methods train dedicated models for each dataset, which means more resources are required when deploying models in realistic situations. To be specific, W\textsuperscript{2}NER and DML-NEZHA-Large are based on BERT-Base (110 million parameters) or BERT-Large (340 million parameters) and include additional layers (e.g., Convolution Layer and Co-Predictor Layer in \citet{li2021unified}). Specifically, if we need to deploy an NER model for $N$ datasets, the number of model parameters is at least $N*340$ million. However, if we use PUnifiedNER, the number of model parameters is equal to the number of T5-base parameters, i.e., 220 million, which is $N*1.55$ times smaller than in W\textsuperscript{2}NER (BERT-Large version) or DML-NEZHA-Large. Also, as reported in previous work \cite{JMLR:v21:20-074,lester-etal-2021-power,lu-etal-2022-unified}, increasing the model size of T5 architecture can further improve performance. Thus, we believe that PUnifiedNER is able to compensate the performance if we change the backbone to T5-Large/XL/XXL without significant increase in the number of parameters needed.

% Finally, our on-demand fine-tuned models ... 

\subsection{Ablation Study} \label{subsec:ablations}

\subsubsection{Ablations on Prefixed Prompts Setups:} 
To verify our prefixed prompt design choices, we perform ablations for different prefixed prompt settings as discussed in the pilot study and report results on five datasets: MSRA, OntoNotes 4.0, Resume, CLUENER and CCKS2021.

The results are shown in Table \ref{tab:ablation_prefixed_design}. We also report overall average performance in the rightmost column. Our default setting is Dataset-Dependent Prompt. We compare this with two ablations: Random Prompt and Random Prompt+Exact Match. We observe that Dataset-Dependent Prompt leads to much better performance compared to Random Prompt (avg. 85.83 vs. 73.24) and Random Prompt+Exact Match (avg. 85.83 vs. 72.31). This makes sense since Exact Match suffers from information leakage in training, thus the performance degrades during inference. While Random Prompt results in too many \textit{``NULL''} examples with few or even zero positive examples where the model learns nothing about ground truth. Dataset-Dependent Prompt is more useful in balancing positive and \textit{``NULL''} examples, reducing the risk of information leakage and improving data efficiency.

\subsubsection{Ablations on Language Model Adaptation Steps:} The pilot study has shown that language model adaptation can largely close the objective gap and boost downstream task performance on the prompting-based seq2seq NER task. In this subsection, we go a step further to verify whether selecting the model with lowest validation loss on language model adaptation can benefit the downstream multi-dataset joint learning. We perform another  multi-dataset joint training with all eight NER datasets based on model after 10k steps and after 50K steps (full training). All training settings are the same as described in the pilot study. Other hyper-parameters settings are presented in the Appendix.\looseness=-1

Table \ref{tab:ablation_lm_steps} shows the results where our default setting of 6K steps surpasses the other two settings: after 10K steps (avg. 84.82 vs. 84.25) and after 50K steps (avg. 84.82 vs. 84.51). This demonstrates that selecting the model with the lowest validation loss is a useful approach. However, other settings outperform the default for some datasets. For example, the model after 50K steps achieves an f-score of 97.24, which is a new state-of-the-art for the Resume dataset. We notice from Figure \ref{fig:prefix_lm} that the model after 50K steps has a high validation loss as compared to the model after 6K steps for the Resume dataset , which suggests that validation loss might not be the only model selection criterion that should be used. We leave this exploration for future work.

\subsection{On-demand Named Entity Recognition}

Our code and a demo interface for PUnifiedNER have been made available,\footnote{All resources are available at: \url{https://github.com/GeorgeLuImmortal/PUnifiedNER}.} which demonstrates the capability of on-demand named entity recognition.

\section{Conclusion}

In this work, we present a novel Prompting-based Unified NER system (PUnifiedNER) that can recognise a large set of entity types in data from various domains and support on-demand entity recognition. To achieve this, we first recast the NER task to a seq2seq task where a prefix language modelling objective is introduced to reduce the gap between pre-training and fine-tuning. A pilot study shows that prefix language modelling is very effective in adapting pre-trained language models. Dataset-Dependent Prompt is designed to unify data from all datasets into a united format that enables joint training crossing multiple datasets for PUnifiedNER. Besides on-demand named entity recognition, experimental results show that the multi-dataset training empowered by prompting can also lead to significant performance gains over single-dataset training, while dramatically reducing model deployment cost. Further, the jointly trained PUnifiedNER model sets a new state-of-the-art performance level for the Resume dataset, and is comparable to other state-of-the-art dataset-specific NER methods in some cases.  Lastly, extensive ablation studies are performed to clarify the design choices of PUnifiedNER.

% \section{Ethical Statement}
% We honor the AAAI Code of Ethics. No private data or non-public information was used in this work.

\section*{Acknowledgements}
We would like to thank \textbf{Hongyu Lin} from Chinese Academy of Sciences for his thoughtful discussion, as well as the many others who have helped. We would also like to thank anonymous reviewers for their insightful comments to help improve the paper. This publication has emanated from research conducted with the support of SenseTime Research.

\bibliography{aaai23}

\appendix

\onecolumn
\section{Appendices}
\section{A Evaluation Metrics}

Regarding evaluation metrics, we follow prior work \cite{yan-etal-2021-unified-generative,li2021unified,lu-etal-2022-unified} and employ macro F1-score. A predicted entity is counted as true positive only if its text span and entity types match those of a gold entity.

\begin{table*}
\small
\centering
\resizebox{\textwidth}{!}{%
\begin{tabular}{lccccccccc}
\hline
\multicolumn{1}{c}{\multirow{2}{*}{\textbf{Dataset}}} & \multicolumn{4}{c}{\textbf{Sentence}}                                &           & \multicolumn{4}{c}{\textbf{Mention}}                                 \\ \cline{2-5} \cline{7-10} 
\multicolumn{1}{c}{}                                  & \textbf{\#All} & \textbf{\#Train} & \textbf{\#Dev} & \textbf{\#Test} & \textbf{} & \textbf{\#All} & \textbf{\#Train} & \textbf{\#Dev} & \textbf{\#Test} \\ \hline
Ecommerce                                             & 4,987          & 3,989            & 500            & 498             &           & 15,216         & 12,109           & 1,540          & 1,567           \\
MSRA                                                  & 50,729         & 44,364           & 4,365              &   -         &           & 80,884         & 74,703           & 6,181              &   -        \\
OntoNotes 4.0                                         & 24,371         & 15,724           & 4,301          & 4,346           &           & 28,006         & 13,372           & 6,950          & 7,684           \\
People Daily 2014                                     & 286,269        & -                & -              & -               &           & 751,229        & -                & -              & -               \\
Boson                                                 & 16,753         & -                & -              & -               &           & 23,172         & -                & -              & -               \\
Resume                                                & 4,759          & 3,819            & 463            & 477             &           & 16,565         & 13,438           & 1,497          & 1,630           \\
CLUENER                                               & 12,091         & 10,748           & 1,343          & 1,350           &           & 27,043         & 23,971           & 3,072          & -               \\
CCKS2021                                              & 10,825         & 8,855            & 1,970          & -               &           & 53,620         & 43,644           & 9,976          & -               \\ \hline
\end{tabular}%
}
\caption{Dataset Statistics. ``\#'' denotes the amount. For MSRA and CCKS2021, we report the result on validation set. For People Daily 2014 and Boson, we sample 2,000 examples and 2,000 examples from all examples for evaluation and testing, respectively. For CLUENER, we submit prediction of test set to the competition and report the score provided by the website.}
\label{tab:data_statistics}
\end{table*}

\begin{table*}
\small
\centering
\resizebox{\textwidth}{!}{%
\begin{tabular}{lcl}
\hline
\multicolumn{1}{c}{\textbf{Dataset}} & \textbf{\#Entity} & \textbf{Entity}                                                                                                                                                                                                                                                                                                               \\ \hline
Ecommerce                            & 2                 & \{'HP(brand)':'品牌','HC(commodity)':'商品'\}                                                                                                                                                                                                                                                                                                       \\\hline
MSRA                                 & 3                 & \{'LOC':'地点','PER':'名称','ORG':'组织'\}                                                                                                                                                                                                                                                                                          \\\hline
OntoNotes 4.0                        & 4                 & \{'GPE':'地缘政治实体','LOC':'地点','PER':'名称','ORG':'组织'\}                                                                                                                                                                                                                                                                           \\\hline
People Daily 2014                    & 4                 & \{'LOC':'地点','PER':'名称','ORG':'组织','T(time)':'时间'\}                                                                                                                                                                                                                                                                                 \\\hline
Boson                                & 6                 & \{'product\_name':'产品','time':'时间', 'person\_name':'名称', 'org\_name':'组织', 'location':'地点', 'company\_name':'公司'\}                                                                                                                                                                                                            \\\hline
Resume                               & 8                 & \{'NAME':'名称', 'CONT':'国籍', 'RACE':'民族', 'TITLE':'职位', 'EDU':'学历', 'ORG':'公司', 'PRO':'专业', 'LOC(place of birth)':'籍贯'\}                                                                                                                                                                                                                       \\\hline
CLUENER                              & 10                & \begin{tabular}[c]{@{}l@{}}\{'name':'名称','company':'公司','game':'游戏','organization':'组织','movie':'电影', 'address':'地点','position':'职位',\\ 'government':'政府','scene':'景点','book':'书籍'\}\end{tabular}                                                                                                                             \\\hline
CCKS2021                             & 17                & \begin{tabular}[c]{@{}l@{}}\{'prov':'省份', 'city':'城市', 'district':'区', 'town':'街道', 'community':'社区', 'poi(point of interest)':'兴趣点', 'road':'路', 'roadno':'路号',\\ 'subpoi':'次兴趣点', 'devzone':'产业园', 'houseno':'楼号', 'intersection':'路口', 'assist':'方位', 'cellno':'单元', 'floorno':'楼层',\\  'distance':'距离', 'village\_group':'村组'\}\end{tabular} \\ \hline
\end{tabular}%
}
\caption{Entity tag of each dataset and the conversion from tag used in dataset to corresponding Chinese natural language. For some tags that are hard to understand, we provide their meaning in brackets. ``\#'' denotes the amount of entity types.}
\label{tab:data_entity_details}
\end{table*}

\section{B Dataset Statistics}

We evaluate our framework on 8 Chinese flat NER datasets. In Table \ref{tab:data_statistics}, we present the detailed statistics. The details of entity type of each dataset is presented in Table \ref{tab:data_entity_details}.

\section{C Implementation Details}

In this section, we provide more details of our experiments. Hyper-parameter settings are listed in Table \ref{tab:hyperparam}. We adopt AdamW \cite{loshchilov2018decoupled} optimizer. Our model is implemented with PyTorch, language model adaptation is trained with 24 NVIDIA 1080Ti GPUs, multi-dataset training is trained with 8 Tesla V100 GPUs. Note that we use beam width equals to 5 in evaluation but 10 in testing. Single-dataset training is trained with one Tesla V100 GPU and we use beam width equals to 10 for both evaluation and testing. Hyper-parameter tuning is based on validation set.

\section{E Multi-dataset Joint Training Performance on Validation Set}

\begin{figure*}
\centering
\includegraphics[width=1.0\textwidth]{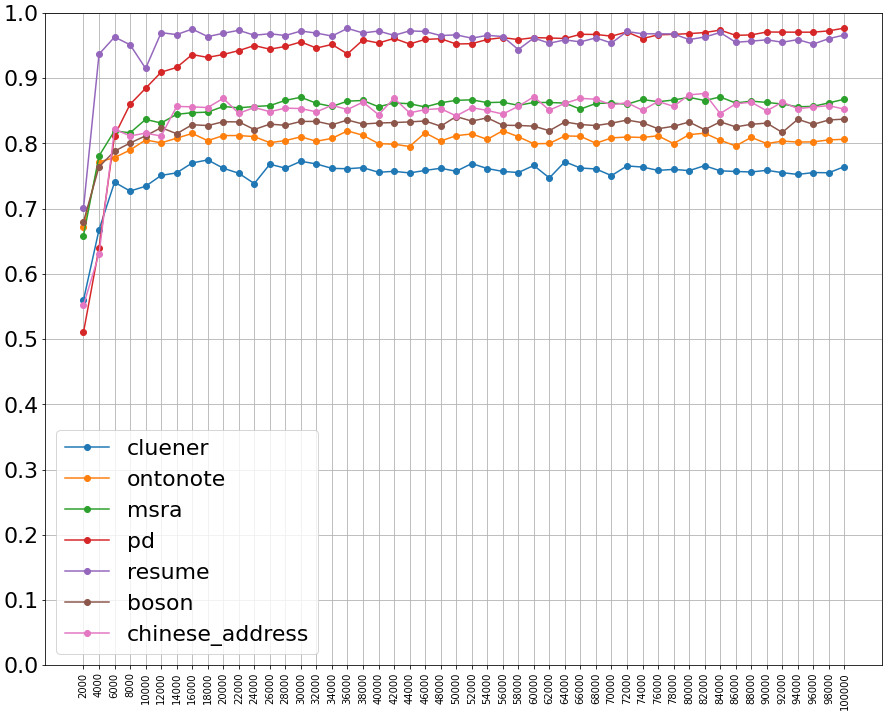}
 \caption{Validation macro f-score of multi-dataset joint training.}
 \label{fig:multi_dev_score}
\end{figure*}

The validation macro f-score of each checkpoint is presented in Figure \ref{fig:multi_dev_score}. We select checkpoint on step 30K since it is generally ideal for all datasets. However, we can observe that after the chosen checkpoint, the validation performance on People Daily 2014 (the red line) consistently increases and reaches the best f-score at step 100K. This explains the suboptimal performance of PUnifiedNER on dataset People Daily 2014.

\begin{table*}
\small
\centering
\resizebox{1.0\textwidth}{!}{%
\begin{tabular}{lcllcllc}
\hline
\multicolumn{2}{c}{\textbf{Language Model Adaptation}}        & \multicolumn{1}{c}{\textbf{}} & \multicolumn{2}{c}{\textbf{Multi-dataset Training}}           & \textbf{} & \multicolumn{2}{c}{\textbf{Single-dataset Training}}          \\ \hline
\textbf{Hyper-parameter} & \multicolumn{1}{l}{\textbf{Value}} & \textbf{}                     & \textbf{Hyper-parameter} & \multicolumn{1}{l}{\textbf{Value}} & \textbf{} & \textbf{Hyper-parameter} & \multicolumn{1}{l}{\textbf{Value}} \\ \hline
Warm-up step             & 1K                                 &                               & Warm-up step             & 5K                                 &           & Warm-up step             & 1K                                 \\
Total step               & 50K                                &                               & Total step               & 100K                               &           & Total epoch              & 40                                 \\
Eval step                & 10K                                &                               & Eval step                & 20K                                &           & Eval epoch               & 1                                  \\
Max input length         & 256                                &                               & Max input length         & 512                                &           & Max input length         & 512                                \\
Max output length        & 256                                &                               & Max output length        & 512                                &           & Max output length        & 512                                \\
Learning rate            & 1e-4                               &                               & Learning rate            & 1e-4                               &           & Learning rate            & 1e-5                               \\
Batch size               & 48                                 &                               & Batch size               & 32                                 &           & Batch size               & 4                                  \\
Beam width               & -                                  &                               & Beam width               & 5                                  &           & Beam width               & 10                                 \\ \hline
\end{tabular}%
}
\caption{Hyper-parameter settings.}
\label{tab:hyperparam}
\end{table*}

\end{CJK}
\end{document}